\documentclass[conference]{IEEEtran}
\IEEEoverridecommandlockouts
\usepackage{cite}
\usepackage{amsmath,amssymb,amsfonts}
\usepackage{algorithmic}
\usepackage{algorithm}
\usepackage{graphicx}
\usepackage{textcomp}
\usepackage{xcolor}
\usepackage[colorlinks]{hyperref}
\usepackage{subcaption}
\usepackage{tikz}
\usepackage{pgfplots}
\pgfplotsset{compat=1.17}
\usepgfplotslibrary{statistics}

\newcommand{\APPFL}{\texttt{APPFL}}
\newcommand{\re}[1]{{\color{black} #1}}

\newtheorem{definition}{Definition}

\def\BibTeX{{\rm B\kern-.05em{\sc i\kern-.025em b}\kern-.08em
    T\kern-.1667em\lower.7ex\hbox{E}\kern-.125emX}}
\begin{document}

\title{APPFL: Open-Source Software Framework for Privacy-Preserving Federated Learning
\thanks{This work was supported by the U.S. Department of Energy, Office of Science, Advanced Scientific Computing Research, under Contract DE-AC02-06CH11357.}
}

\author{
\IEEEauthorblockN{Minseok Ryu}
\IEEEauthorblockA{\textit{Mathematics and Computer Science Division} \\
\textit{Argonne National Laboratory}\\
Lemont, IL, USA \\
mryu@anl.gov}
\\
\IEEEauthorblockN{Kibaek Kim}
\IEEEauthorblockA{\textit{Mathematics and Computer Science Division} \\
\textit{Argonne National Laboratory}\\
Lemont, IL, USA \\
kimk@anl.gov}
\and
\IEEEauthorblockN{Youngdae Kim}
\IEEEauthorblockA{\textit{Mathematics and Computer Science Division} \\
\textit{Argonne National Laboratory}\\
Lemont, IL, USA \\
youngdae@anl.gov}
\\
\IEEEauthorblockN{Ravi K. Madduri}
\IEEEauthorblockA{\textit{Data Science and Learning
Division} \\
\textit{Argonne National Laboratory}\\
Lemont, IL, USA \\
madduri@anl.gov}
}

\maketitle

\begin{abstract}
Federated learning (FL) enables training models at different sites and updating the weights from the training instead of transferring data to a central location and training as in classical machine learning.
The FL capability is especially important to domains such as biomedicine and smart grid, where data may not be shared freely or stored at a central location because of policy regulations. 
Thanks to the capability of learning from decentralized datasets, FL is now a rapidly growing research field, and numerous FL frameworks have been developed.
In this work we introduce \APPFL{}, the Argonne Privacy-Preserving Federated Learning framework. \APPFL{}  allows users to leverage implemented privacy-preserving algorithms, implement new algorithms, and simulate and deploy various FL algorithms with privacy-preserving techniques. The modular framework enables users to customize the components for algorithms, privacy, communication protocols, neural network models, and user data.
We also present a new communication-efficient algorithm based on an inexact alternating direction method of multipliers. The algorithm requires significantly less communication between the server and the clients than does the current state of the art.
We demonstrate the computational capabilities of \APPFL{}, including differentially private FL on various test datasets and its scalability, by using multiple algorithms and datasets on different computing environments.
\end{abstract}


\section{Introduction}




Federated learning (FL) is a growing research field in machine learning (ML). FL enables multiple institutions (or devices) to collaboratively learn without sharing data.
Specifically, FL is a form of distributed learning with the goal of training a global ML model by systematically updating weights from training on local and decentralized data.
Partly because of its learning capability without sharing data, FL is listed as one of the key technologies to address the U.S. Department of Energy's and National Institutes of Health's grand challenges on adopting artificial intelligence (AI) to tackle complex biomedical data (e.g., Bridge2AI program \cite{Bridge2AI}).
Moreover, recent reports have highlighted the increasing need to enable AI/ML for privacy-sensitive datasets (e.g., Chapter 15 of~\cite{schmidt2021national}).

FL by itself, however, does not guarantee the privacy of data, because the information extracted from the communication of FL algorithms can be accumulated and effectively utilized to infer the private local data used for training (e.g.,~\cite{wei2020framework,ryu2021privacy,ryu2021differentially,ryu2022differentially}).
In addition to data privacy, challenges exist in FL in areas of algorithm design, statistics, and software architecture. Privacy-preserving techniques have been studied and integrated into FL algorithms (e.g.,~\cite{ryu2021differentially,ryu2022differentially,choudhury2019differential,kang2020federated}), often named as privacy-preserving FL (PPFL). 
While FL algorithms have been advanced to achieve greater accuracy and scalability, practical use and integration of new techniques such as privacy preservation require much more careful algorithm design to ensure efficient communication (see, e.g.,~\cite{zhou2021communication,chen2021communication}). 
For increased adoption of FL techniques, research communities need to develop not only new PPFL open-source frameworks but also new benchmarks using existing implementations.

In addition to the streamlined deployment of PPFL packages, simulation of PPFL is particularly important for quantification of model performance, learning, and privacy preservation. These simulations are compute intensive and challenging to scale with the increasing number of FL clients and differences in the sample size at each client. 
Therefore, a scalable simulation capability is necessary for PPFL packages.

In this paper we introduce the Argonne Privacy-Preserving Federated Learning (\texttt{APPFL}) framework, an open-source PPFL framework that (i) provides application programming interfaces to easily implement and combine key algorithmic components required for PPFL in a plug-and-play manner; (ii) can be used for simulations on high-performance computing (HPC) architecture with MPI; and (iii) runs on heterogeneous architectures. Examples of algorithmic components include FL algorithms, privacy techniques, communication protocols, FL models to train, and data.
We present a new communication-efficient FL algorithm based on an inexact alternating direction method of multipliers (IADMM), which significantly reduces the data that is needed to iteratively communicate between the server and clients, as compared with the inexact communication-efficient ADMM (ICEADMM) algorithm recently developed in~\cite{zhou2021communication}. In \APPFL{}, in addition to the well-known federated averaging (FedAvg) algorithm~\cite{mcmahan2017communication} as a special case of IADMM algorithms, we have implemented our new algorithm, as well as the ICEADMM algorithm. 

We also present the performance results from the \texttt{APPFL} framework with two communication protocols: remote procedure calls (gRPC \cite{wang1993grpc}) and the Message Passing Interface (MPI). While the quantitative results from using gRPC mimic those from using PPFL on heterogeneous computing architectures, MPI allows scalable simulations of PPFL by utilizing multiple GPUs on high-performance computing architecture. In our demonstration we report and discuss (i) the strong-scaling results from \texttt{APPFL} on the Summit supercomputer at Oak Ridge National Laboratory and (ii) communication efficiency with gRPC in  practical settings. We also discuss the implications of using gRPC on 
heterogeneous computing machines.

We summarize our contributions as follows.
\begin{enumerate}
    \item We develop an open-source software package \texttt{APPFL}, a PPFL framework that provides various capabilities needed to implement and simulate PPFL.
    \item We develop a new communication-efficient IADMM algorithm that significantly reduces the communication as compared with the ICEADMM algorithm~\cite{zhou2021communication}.
    \item We provide extensive numerical results by simulating PPFL with gRPC and MPI with respect to the performance of PPFL algorithms and communication.
\end{enumerate}
The rest of the paper is organized as follows.
We present the architecture of \texttt{APPFL}, PPFL algorithms implemented in \texttt{APPFL}, and numerical demonstration of \texttt{APPFL} in Sections \ref{sec:architecture}, \ref{sec:algorithms}, and \ref{sec:demo}, respectively.
In Section \ref{sec:conclusion} we summarize our conclusions and discuss future implementations to enhance the capability of \texttt{APPFL}.

\section{APPFL Architecture} \label{sec:architecture}
\texttt{APPFL} is an open-source Python package that provides privacy-preserving federated learning tools for users in practice while allowing research communities to implement, test, and validate various ideas for PPFL.
The source code is available in a public GitHub repository, and \APPFL{} v0.0.1~\cite{appfl0.0.1} has been released as a package distributed and can be easily installed via pip \url{https://pypi.org/project/appfl/}.
In this section we present an overview of the \texttt{APPFL} architecture and compare \texttt{APPFL} with several existing FL frameworks.


\subsection{Overview}

\begin{figure*}[!tb]
    \centering
    \includegraphics[scale=.5]{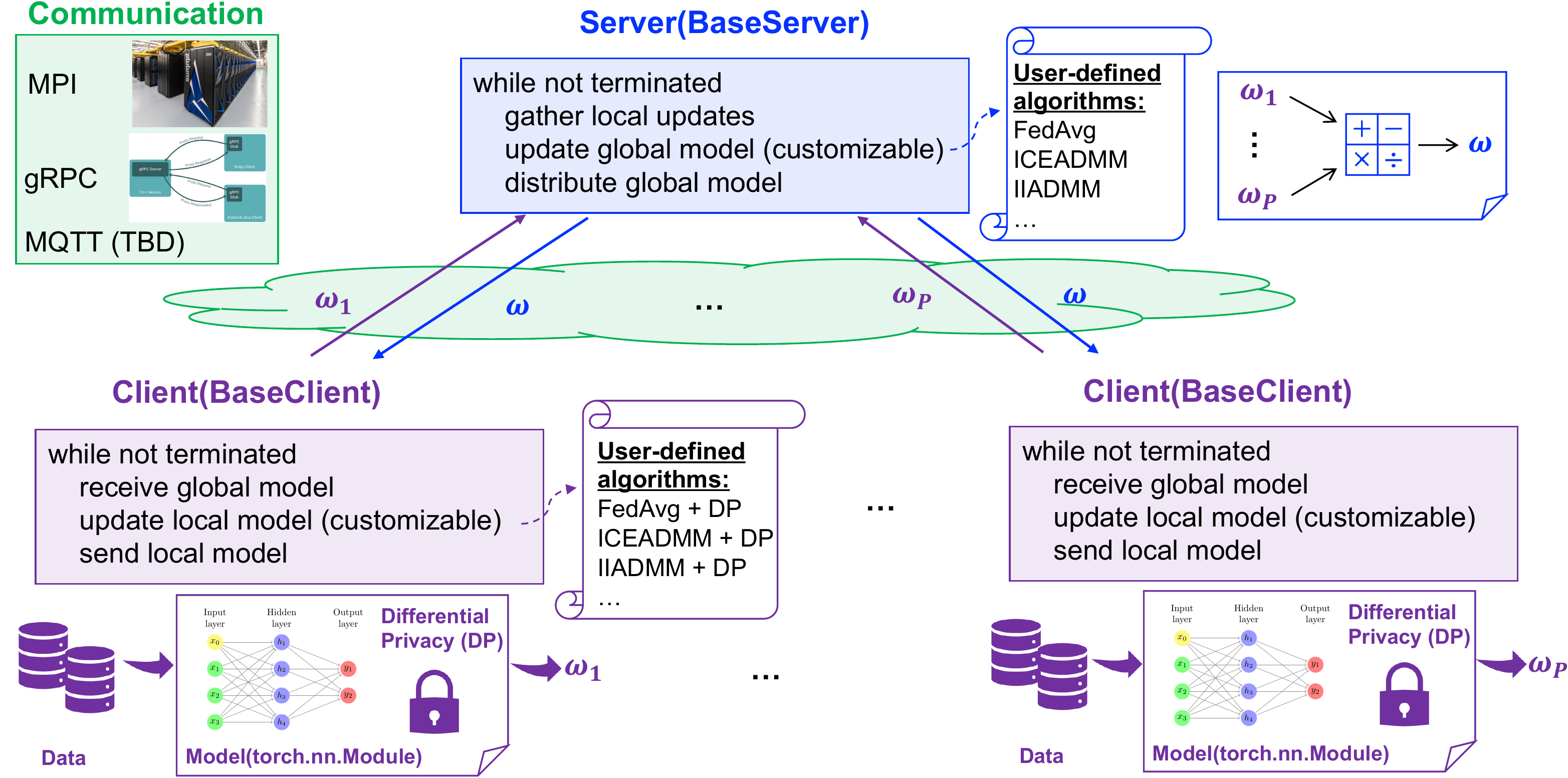}
    \caption{Overview of APPFL architecture}
    \label{fig:appfl}
\end{figure*}

Figure~\ref{fig:appfl} presents an overview of the \APPFL{} architecture.
The \APPFL{} framework has five major components:
(i) federated learning algorithms,
(ii) differential privacy schemes,
(iii) communication protocols,
(iv) neural network models, and 
(v) data for training and testing.

\subsubsection{FL algorithm}

A federated learning algorithm determines how a server updates global model parameters based on local model parameters trained and sent by clients.
Our framework assumes the following general structure of FL:
\begin{align}
    \min_{w} \sum_{p=1}^P \Big\{ \frac{I_p}{I} \sum_{i \in \mathcal{I}_p} \frac{1}{I_p} f_i(w; x_i, y_i) \Big\},  \label{FLModel}        
\end{align}
where
$w \in \mathbb{R}^m$ is a model parameter and 
$x_i$ and $y_i$ are the $i$th data input and data label, respectively,  
$P$ is the number of clients,
$\mathcal{I}_p$ is an index set of data points from a client $p$,
$I_p := | \mathcal{I}_p|$ is the total number of data points for a client $p$, 
$I = \sum_{p=1}^P I_p$ is the total number of data points,
and $f_i$ is the loss of the prediction on $(x_i, y_i)$ made with $w$.
The objective function $f_i$ can be convex (e.g., linear model) and nonconvex (e.g., neural network model).

In \APPFL{}, we currently have implemented the popular FedAvg algorithm~\cite{mcmahan2017communication} and two IADMM-based algorithms: our new improved training- and communication-efficient IADMM (IIADMM) algorithm (see Section~\ref{sec:IIADMM}) and the ICEADMM algorithm~\cite{zhou2021communication}.
Additional user-defined FL algorithms can be implemented by inheriting our Python class \texttt{BaseServer} and implementing the virtual function \texttt{update()}.
For IADMM-based algorithms, clients may need to perform additional work during training, such as forming an augmented Lagrangian function and updating dual variables.
This additional work can be customized as well by inheriting our \texttt{BaseClient} class and implementing the virtual function \texttt{update()}.

\subsubsection{Differential privacy}

Differential privacy schemes enable protecting data privacy by making it difficult to deduce confidential information from aggregated data such as a local model parameter $\omega_i$ of client $i$'s.
The work \cite{geiping2020attack} shows that one can recover an original image with high accuracy using only gradients sent to the server, without sharing the training data between clients and a server.
Therefore, additional privacy-preserving schemes such as differential privacy capability are critical for a privacy-preserving FL.
We currently support the output perturbation method based on the Laplace mechanism \cite{dwork2014algorithmic}, namely, adding Laplacian noise to the local model parameters before sending them to the server. We plan to add more advanced schemes in the near future.
More details are given in ~\ref{sec:DP}.
Users can also implement their own differential privacy schemes by implementing them in the virtual function \texttt{update()} of \texttt{BaseClient} class.

\subsubsection{Communication protocol}

Two communication protocols, MPI and gRPC, have been implemented in our \APPFL{} framework for sharing model parameters between clients and a server.
The MPI protocol provides an efficient communication method in a cluster environment by utilizing collective communication and remote direct memory access capabilities.
On the other hand, gRPC enables communication between multiple platforms and languages, which will likely be the case for cross-silo FL.
For an efficient cross-device FL involving a massive number of devices (e.g.,~\cite{beckman2016waggle}), we plan to support MQTT, a lightweight, publish-subscribe network protocol that transports messages between devices (\url{https://mqtt.org/}).

\subsubsection{User-defined model}

The user-defined model is a neural network model that inherits PyTorch's neural network module (\texttt{torch.nn.Module}).
All clients are supposed to use the same neural network model architecture for training and testing.
\APPFL{} does not assume anything about the model other than that it inherits \texttt{torch.nn.Module}; users can freely specify their own neural network models.

\subsubsection{User-defined data}

Each client is required to define the training data, which is typically not accessible from the server or any other clients.
We leverage the concept of the PyTorch dataset in \APPFL{}. Users can load their datasets to our \APPFL{} framework by using the \texttt{Dataset} class that inherits the PyTorch \texttt{Dataset} class. This allows us to utilize the PyTorch's \texttt{DataLoader} that provides numerous useful functions including data shuffling and mini-batch training.
When testing data is available at a server, \APPFL{} provides a validation routine that evaluates the accuracy of the current global model.
This validation can be used to monitor and determine the convergence of an FL.

\subsection{Existing FL frameworks}
A few open-source FL frameworks exist. 
These include Open Federated Learning (\texttt{OpenFL})  \cite{reina2021openfl}, Federated Machine Learning (\texttt{FedML}) \cite{he2020fedml}, TensorFlow Federated (\texttt{TFF}) \cite{TFF}, and \texttt{PySyft} \cite{ziller2021pysyft}. 
In Table \ref{Table:comparison} we compare them based on advanced functionality available in \texttt{APPFL}.
See \cite{kholod2021open} for a more detailed summary and comparison of the existing open-source FL frameworks.

\begin{table}[!htb]
\centering    
\caption{Comparison of \texttt{APPFL} with some of the existing open-source FL frameworks}
\label{Table:comparison}
\begin{tabular}{|c|c|c|c|c||c|}
\hline
    & \texttt{OpenFL} & \texttt{FedML}  & \texttt{TFF} & \texttt{PySyft} & \texttt{APPFL}  \\  \hline
Data privacy &  &  & \checkmark & \checkmark & \checkmark \\
MPI  &   & \checkmark  &   &   & \checkmark\\
gRPC & \checkmark &  \checkmark &    &  & \checkmark \\
MQTT &   & \checkmark &   &   &   \\    
\hline            
\end{tabular}    
\end{table}

Here we briefly discuss the capabilities of each framework in terms of their relevance to \texttt{APPFL}.

\subsubsection{OpenFL}

This is an open-source FL framework developed by Intel.
It was initially developed as part of a research project on FL for healthcare and designed for a multi-institutional setting.
In \texttt{OpenFL}, an FL environment is constructed based on collaborator and aggregator nodes that form a star topology; in other words, all collaborator nodes are connected to an aggregator node.
Communication between nodes is through gRPC via a mutually authenticated transport layer security network connection.

\subsubsection{FedML}

This is an open research library to facilitate FL algorithm development and fair performance comparison.
It supports on-device training for edge devices, distributed computing, and single-machine simulation.
It utilizes gRPC and MQTT for device communication to simulate cross-device FL on real-world hardware platforms.
Also, it utilizes MPI for simulating FL in a distributed-computing setting.
Regarding the privacy and security aspect, it implements weak differential privacy that aims to prevent a backdoor attack, which requires less noise in training data compared with what is required for ensuring data privacy \cite{sun2019can}.

\subsubsection{Tensor Flow Federated (TFF)}

This is an open-source framework from Google for machine learning and other computations on decentralized data \cite{TFF}.
In \texttt{TFF}, an FL environment is constructed by using multiple GPUs that are used as clients.
Also, \texttt{TFF} can be simulated on a Google Cloud platform.
Currently, \texttt{TFF} supports FedAvg and differential privacy for private federated learning.

\subsubsection{PySyft}

This is an open-source FL framework from OpenMined, an open-source community \cite{ziller2021pysyft}.
In \texttt{PySyft}, an FL environment is constructed by  \textit{Virtual Workers}, \textit{WebSocket Workers}, or \textit{GridNodes}.
While Virtual Workers live on the same machine and do not communicate over the network, the others leverage WebSocket as a communication medium to ensure that a broad range of devices can participate in a \texttt{PySyft} network.
Currently, \texttt{PySyft} supports FedAvg and differential privacy for private federated learning.

\section{Privacy-Preserving Algorithms} \label{sec:algorithms}
In this section we present our new communication-efficient algorithm, IIADMM. This algorithm  significantly reduces the amount of information transfer between the server and the clients, as compared with the ICEADMM algorithm recently developed in~\cite{zhou2021communication}. 
In Section \ref{sec:IIADMM} we also show that FedAvg \cite{mcmahan2017communication}, the popular FL algorithm, is a special form of the IADMM-based algorithms.
Furthermore, we describe differential privacy (DP) techniques applied to the FL algorithms in Section \ref{sec:DP}.

\subsection{IIADMM} \label{sec:IIADMM}
We consider the reformulation of \eqref{FLModel} given as follows:
\begin{subequations}
\label{FLModel_primal}    
\begin{align}
    \min_{w, \{z_p\}_{p=1}^P} \ & \sum_{p=1}^P  \Big\{ \frac{I_p}{I}  \sum_{i \in \mathcal{I}_p} \frac{1}{I_p}  f_{i} (z_p; x_{i}, y_{i}) \Big\} \\
\mbox{s.t.} \ & w=z_p, \forall p \in [P], \label{FLModel_primal-1}
\end{align}
\end{subequations}
where 
$w \in \mathbb{R}^m$ is a global model parameter and
$z_p \in \mathbb{R}^m$ is a local model parameter defined for every client $p \in [P]:=\{1,\dots,P\}$.
The Lagrangian dual formulation of \eqref{FLModel_primal} is given by
\begin{align*}
\max_{ \{\lambda_p\}_{p=1}^P} \min_{w, \{z_p\}_{p=1}^P } \ \sum_{p=1}^P \Big\{ \frac{1}{I} \sum_{i \in \mathcal{I}_p} f_{i} (z_p; x_{i}, y_{i})  + \langle \lambda_p, w - z_p \rangle \Big\},    
\end{align*}
where $\lambda_p \in \mathbb{R}^m$ is a dual vector associated with the consensus constraints \eqref{FLModel_primal-1}.
Then, ADMM steps \cite{boyd2011distributed} are given by
\begin{subequations}
\begin{align}
& w^{t+1} \leftarrow \text{arg} \min_{w \in \mathbb{R}^m} \ \sum_{p=1}^P \Big( \langle \lambda^t_p, w \rangle + \frac{\rho^t}{2} \|w - z^t_p\|^2 \Big), \label{ADMM-1} \\
& z^{t+1}_p \leftarrow \text{arg} \min_{z_p \in \mathbb{R}^m} \ \frac{1}{I} \sum_{i \in \mathcal{I}_p}  f_{i} (z_p; x_{i}, y_{i}) - \langle \lambda^t_p, z_p \rangle \nonumber \\ 
& \hspace{25mm} +  \frac{\rho^t}{2} \|w^{t+1}-z_p\|^2, \ \forall p \in [P], \label{ADMM-2} \\
& \lambda^{t+1}_p \leftarrow  \lambda^{t}_p + \rho^t (w^{t+1}-z^{t+1}_p), \ \forall p \in [P], \label{ADMM-3}
\end{align}
\end{subequations}
where $\rho^t > 0$ is a hyperparameter, the choice of which may be sensitive to the learning performance, similar to the learning rate of the stochastic gradient descent (SGD) method.
\re{In the context of FL, the global model parameter $w^{t+1}$ in \eqref{ADMM-1} is updated at the central server based on the local model parameters $z_p^t$ and $\lambda_p^t$, where $z_p^t$ is a local primal and $\lambda_p^t$ is a local dual information.  
For every clients $p$, the local parameters $z_p^t$ are updated at the clients by using the global model parameter $w^{t+1}$ given from the server, as in \eqref{ADMM-2} and \eqref{ADMM-3}, respectively. In ADMM, both local primal and dual information $(z_p^t, \lambda_p^t)$ are sent from each client $p$ to the central server is required.}

\re{To reduce the computation burden of \eqref{ADMM-2} without affecting the overall convergence,} the subproblem \eqref{ADMM-2} can be replaced with its inexact version:
\begin{align}
& z^{t+1}_p \leftarrow  \text{arg} \min_{z_p \in \mathbb{R}^m} \ \langle g(z_p^t), z_p \rangle - \langle \lambda^t_p, z_p \rangle \nonumber \\ 
& \hspace{25mm} +  \frac{\rho^t}{2} \|w^{t+1}-z_p\|^2 + \frac{\zeta^t}{2} \|z_p-z^t_p\|^2, \label{IADMM-2}
\end{align}
where $g(z_p^t) =\frac{1}{I} \sum_{i \in \mathcal{I}_p} \nabla f_i (z_p^t)$ is a gradient at $z_p^t$
\re{, and $\zeta^t$ is a proximity parameter that controls the distance between the new iterate $z_p^{t+1}$ and the previous iterate $z_p^t$.}

A process $ \{ \eqref{ADMM-1} \rightarrow \eqref{IADMM-2} \rightarrow \eqref{ADMM-3} \}_{t=1}^T$ is denoted by IADMM.
We improve the IADMM process by conducting 
(i) multiple local primal updates using batches of data, namely, iteratively solving \eqref{IADMM-2} based on batches of data $\mathcal{I}_p$, and
(ii) two independent but identical dual updates at both server and clients, which result in eliminating the need for communicating dual information between clients and the server.
We refer to the proposed algorithm as IIADMM and present its  steps in Algorithm \ref{algo:IIADMM}.

\begin{algorithm}
    \caption{IIADMM}
    \label{algo:IIADMM}
    \begin{algorithmic}[1]
    \STATE Initialize $z^1_1, \ldots, z^1_P, \lambda^1_1, \ldots, \lambda^1_P \in \mathbb{R}^{m}$
    \FOR{each round $t=1,2,\dots,T$}  
        \STATE $w^{t+1} \leftarrow \frac{1}{P} \sum_{p=1}^P (z_p^t - \frac{1}{\rho^t} \lambda^t_p )$
        \FOR{each agent $p \in [P]$ in parallel}
            \STATE $z_p^{t+1} \gets$ \textbf{ClientUpdate}$(t, p, w^{t+1}; z^1_p, \lambda^1_p)$
            \STATE $\lambda^{t+1}_p \leftarrow \lambda^t_p + \rho^t(w^{t+1}-z^{t+1}_p)$
        \ENDFOR    
    \ENDFOR
    \STATE \nonumber
    \STATE \textbf{ClientUpdate}$(t, p, w^{t+1}; z^1_p, \lambda^1_p)$
    \STATE Initialize $z^{1,1} \gets w^{t+1}$
    \STATE Split $\mathcal{I}_p$ into a collection $\{\mathcal{B}_1, \ldots, \mathcal{B}_{B_p} \}$ of batches
    \FOR{local step $\ell=1,\dots,L$}
        \FOR{batch $b = 1, \dots, B_p$}    
        \STATE gradient: $g \gets \frac{1}{|\mathcal{B}_b|}  \sum_{i \in \mathcal{B}_b} \nabla f_i(z^{\ell,b}) $        
        \STATE update: 
        \begin{align}
        z^{\ell,b+1} \leftarrow & z^{\ell,b}- \frac{g - \lambda_p^t  - \rho^t( w^{t+1} - z^{\ell,b} )}{\rho^t + \zeta^t}
        \end{align}
        \STATE if $b=B_p$: $z^{\ell+1,1} \gets z^{\ell, B_p}$
        \ENDFOR
    \ENDFOR
    \STATE $z^{t+1}_p \gets z^{L+1,1}$
    \STATE $\lambda^{t+1}_p \leftarrow \lambda^t_p + \rho^t(w^{t+1}-z^{t+1}_p)$
    \RETURN $z^{t+1}_p$
    \end{algorithmic}
\end{algorithm}

Specifically,
in line 3, the global model parameter $w^{t+1}$ is updated based on a closed-form solution expression of \eqref{ADMM-1}.
\re{This global model update is conducted at the central server (lines 1--8). The resulting global model parameter is distributed to all clients in line 5. Then,}
the local model parameters $\{z^{t+1}_p\}_{p=1}^P$ are updated at each client side through the multiple local primal updates \re{using the batches of data} described in lines 10--22.
\re{Specifically, the global model parameter $w^{t+1}$ received from the central server is set to be an initial point $z^{1,1}$ for local model updates in line 11.
In line 12, the local data is split to several batches.
For every local step $\ell$ and batch $b$, the gradient of the loss function is computed in line 15, and the local model parameter is updated based on a closed-form solution expression of \eqref{IADMM-2} in line 16.
After updating the local primal parameters $z_p^{t+1}$ in line 20, the dual parameters $\lambda_p^t$ is updated via \eqref{ADMM-3} in line 21. Then only the local primal parameters are sent to the central server, i.e., from line 22 to line 5.
In line 6, the dual parameters $\lambda^t_p$ is updated via \eqref{ADMM-3}.
}
Note that the two independent dual updates in line 21 and line 6 are identical for every round because the initial local primal and dual information $(z^1, \lambda^1)$ is shared once at the beginning of the algorithm.

The proposed IIADMM is similar to ICEADMM \cite{zhou2021communication} in that both are variants of IADMM.
However, ICEADMM conducts multiple local primal and dual updates without using the batches of data, namely, iteratively solving \eqref{IADMM-2} and \eqref{ADMM-3} for $L$ times while $B_p = 1$.
This method of local updates results in communicating not only primal but also dual information from clients to the server for every communication round, which can be a significant communication burden particularly in an FL setting, as discussed in~\ref{sec:gRPC}.
Nevertheless, a benefit of utilizing the dual information is a potential improvement on the performance of the algorithm, for example, by introducing an adaptive penalty, as discussed in \cite{xu2017adaptive} and \cite{mhanna2018adaptive}.

We also highlight that IADMM is a generalization of the well-known FedAvg \cite{mcmahan2017communication} composed of 
(i) averaging local model parameters for a global update, namely, $w^{t+1} = \frac{1}{P}\sum_{p=1}^P z_p^{t}$, and
(ii) SGD steps for local updates, namely, $z_p^{t+1} = z_p^t - \eta g(z_p^t)$, where $\eta$ is a learning rate (or step size) because FedAvg utilizes the primal information (i.e., $z_p^t$) for updating a global model parameter, while IADMM utilizes not only primal but also dual information (i.e., $\lambda^t_p$).
One  can  easily see from ICEADMM \cite{zhou2021communication} that FedAvg is a special case of ICEADMM by setting $\lambda^t = 0$, $\zeta^t=0$, $\rho^t = 1/\eta$ for every iteration $t$. 

To summarize, the proposed IIADMM utilizes both primal and dual information for updating a global parameter and has the potential to improve learning performance by utilizing the dual information (i.e., a benefit from ICEADMM) while communicating only primal information (i.e., a benefit from FedAvg).

\subsection{Differential privacy} \label{sec:DP}
In \texttt{APPFL}, DP techniques are integrated with the FL algorithms for learning while preserving data privacy against an inference attack \cite{shokri2017membership} that can take place in any communication round. 

\begin{definition}
A randomized function $\mathcal{A}$ provides $\bar{\epsilon}$-DP if, 
for any two datasets $\mathcal{D}$ and $\mathcal{D}'$ that differ in a single entry and for any set $\mathcal{S}$,
\begin{align}
\Big| \ln \Big( \frac{\mathbb{P}( \mathcal{A}(\mathcal{D}) \in \mathcal{S} ) }{\mathbb{P}( \mathcal{A}(\mathcal{D}') \in \mathcal{S} )  } \Big) \Big| \leq  \bar{\epsilon},    \label{DP_def}
\end{align}
where $\mathcal{A}(\mathcal{D})$ (resp. $\mathcal{A}(\mathcal{D'})$) is a randomized output of $\mathcal{A}$ on input $\mathcal{D}$ (resp. $\mathcal{D}'$).
\end{definition}

This implies that as $\bar{\epsilon}$ decreases, it becomes hard to distinguish the two datasets $\mathcal{D}$ and $\mathcal{D}'$ by analyzing the randomized output of $\mathcal{A}$.
Here, $\bar{\epsilon}$ is a privacy budget indicating that stronger privacy is achieved with a lower $\bar{\epsilon}$. 

A popular way of constructing a randomized function $\mathcal{A}$ that ensures $\bar{\epsilon}$-DP is to add some noise directly to the true output $\mathcal{T}(\mathcal{D})$, namely, 
\begin{align}
\mathcal{A}(\mathcal{D}) = \mathcal{T}(\mathcal{D}) + \tilde{\xi},  
\end{align}
which is known as the output perturbation method. 
Several types of noise $\tilde{\xi}$  lead to $\bar{\epsilon}$-DP.
An example is Laplacian noise extracted from a Laplace distribution with zero mean and scale parameter $b := \bar{\Delta} / \bar{\epsilon}$, where $\bar{\epsilon}$ is from \eqref{DP_def} and  $\bar{\Delta} \geq \max_{\mathcal{D}' \in \mathcal{N}(\mathcal{D}) } \| \mathcal{A}(\mathcal{D}) - \mathcal{A}(\mathcal{D}') \|$ is an upper bound on the sensitivity of the output with respect to the collection $\mathcal{N}(\mathcal{D})$ of datasets $\mathcal{D}'$ differing in a single entry from the given dataset $\mathcal{D}$ \cite{dwork2014algorithmic}.

In Section~\ref{sec:demo} we demonstrate the Laplace-based output perturbation method that guarantees $\bar{\epsilon}$-DP on data for any communication round of the FL algorithms implemented in \texttt{APPFL}.
In the output perturbation method, the true output (i.e., $z_p^{t+1}$ in line 20 in Algorithm \ref{algo:IIADMM}) is perturbed by adding the noise $\tilde{\xi}$ generated by a Laplace distribution with zero mean and the scale parameter $b=\bar{\Delta}/\bar{\epsilon}$, where $\bar{\Delta}$ should satisfy $\bar{\Delta} \geq \frac{1}{\rho^t+\zeta^t} \max_{\mathcal{D}' \in \mathcal{N}(\mathcal{D})} \|g(\mathcal{D}) - g(\mathcal{D}') \|$.
Clipping the gradient by a positive constant $C$ leads to $\|g\| \leq C$, which allows us to set $\bar{\Delta}=2C/({\rho^t+\zeta^t})$.
After the perturbation, the resulting randomized outputs are sent to the server.

More advanced methods exist that guarantee DP other than the output perturbation. 
For example, an objective perturbation method \cite{chaudhuri2011differentially} ensures DP on data by perturbing the objective function of an optimization problem rather than perturbing the output of the problem.
As theoretically shown in \cite{chaudhuri2011differentially, zhang2016dynamic}, the objective perturbation provides more accurate learning than does the output perturbation. 
As a future implementation of \texttt{APPFL}, we plan to incorporate advanced DP methods for improving the performance of the PPFL algorithms.

\section{Demonstration of APPFL} \label{sec:demo}

In this section we demonstrate the capabilities of \texttt{APPFL} by extensive experimentation using test datasets on different computing architectures.
For all experiments, we use \APPFL{} version 0.0.1, available through pip. The code for this demonstration is also available at \url{https://github.com/APPFL/APPFL}.

\begin{figure*}[!htp]
    \centering
    \begin{subfigure}[b]{0.24\textwidth}
        \centering      
        \includegraphics[width=\textwidth]{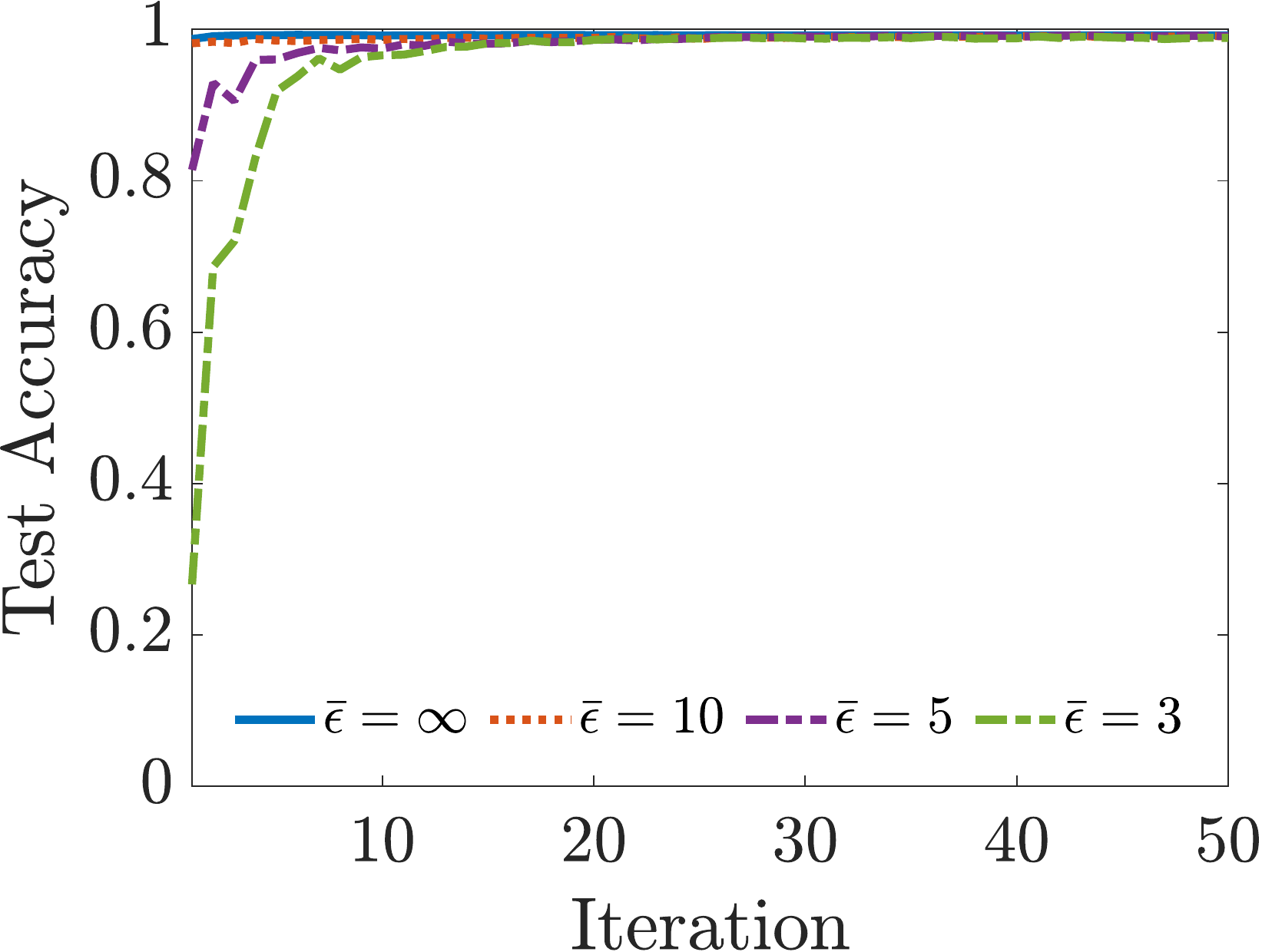}
        \includegraphics[width=\textwidth]{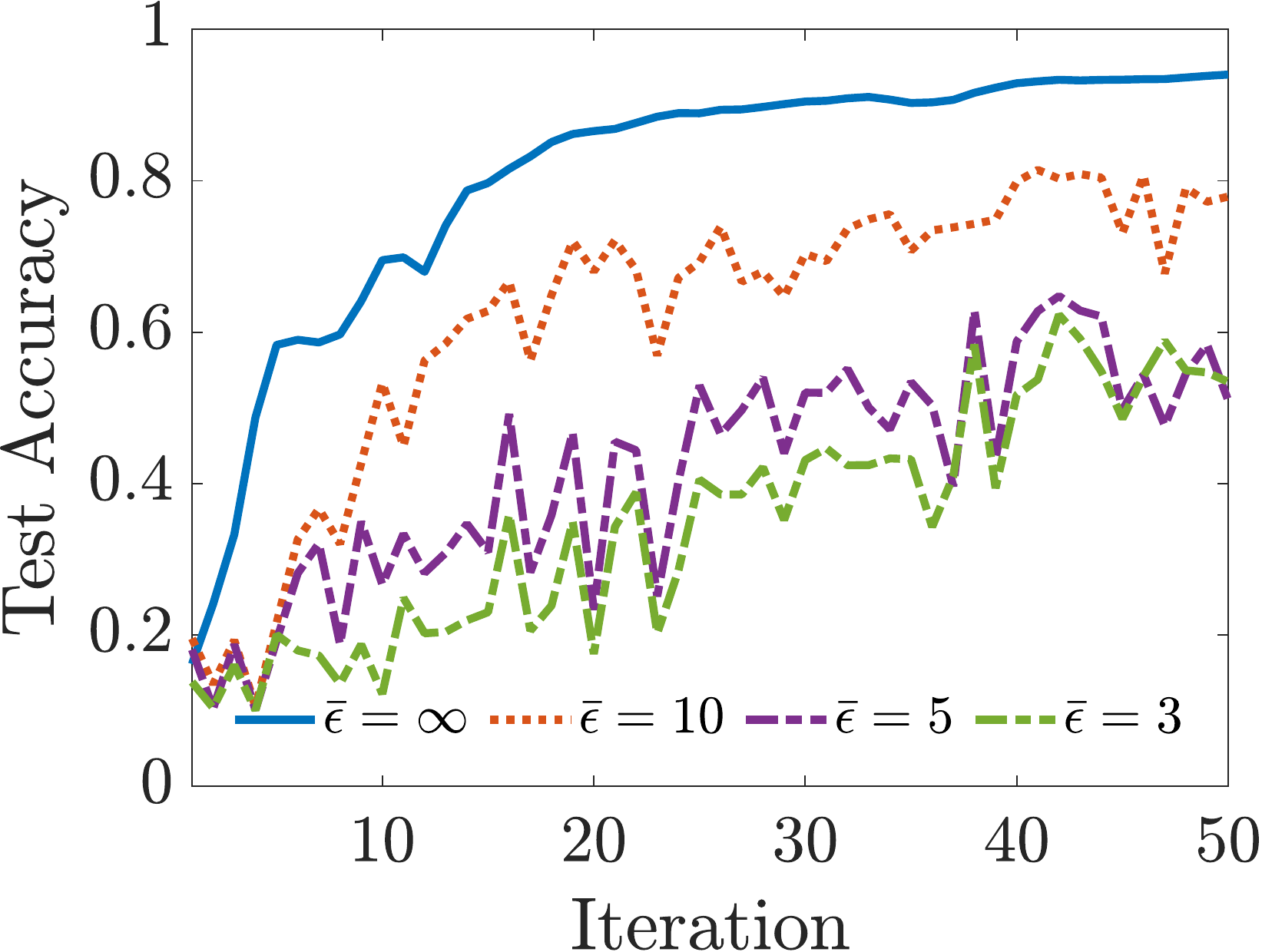}
        \includegraphics[width=\textwidth]{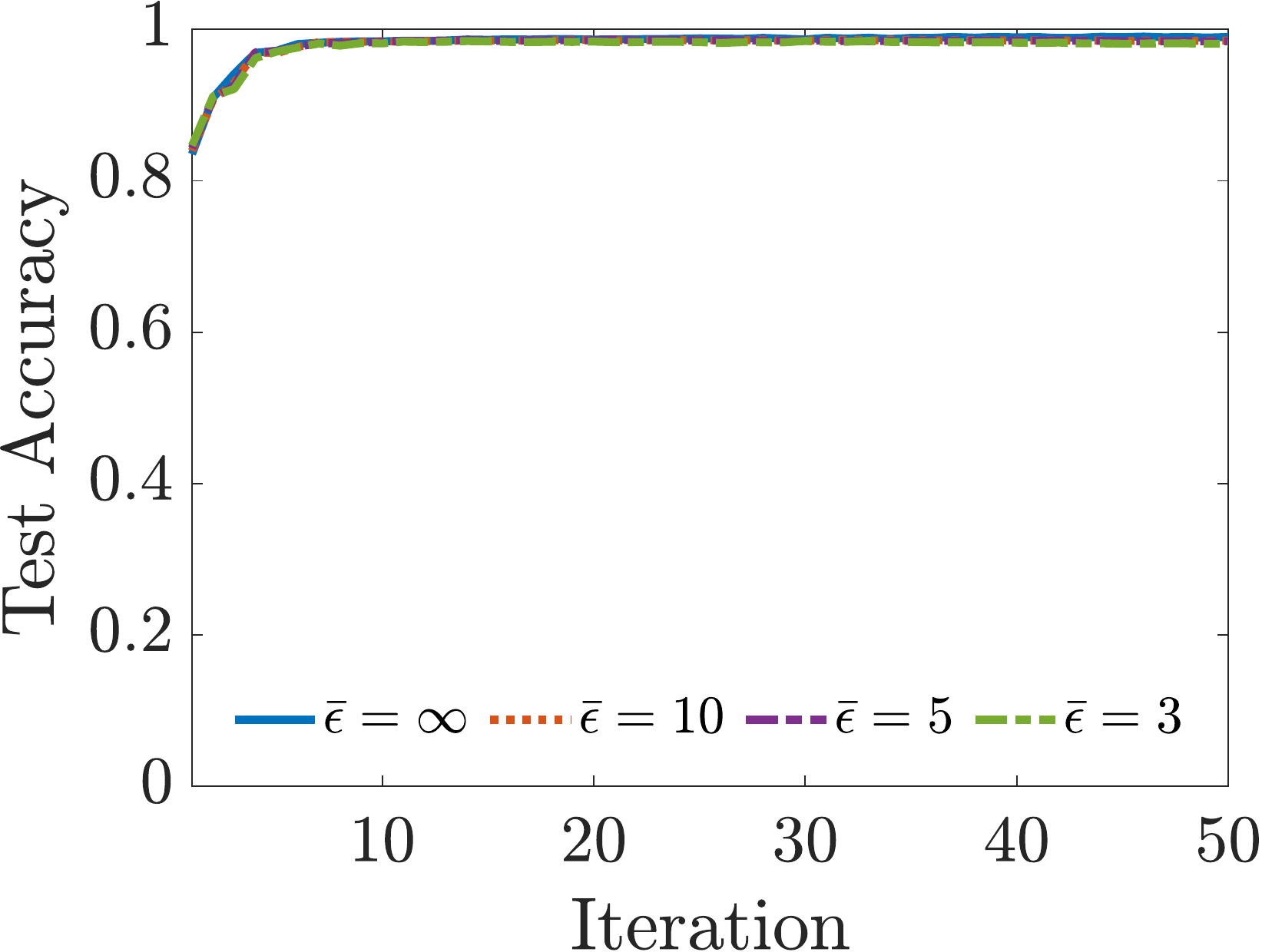}
        \caption{MNIST}
    \end{subfigure}
    \begin{subfigure}[b]{0.24\textwidth}
        \centering      
        \includegraphics[width=\textwidth]{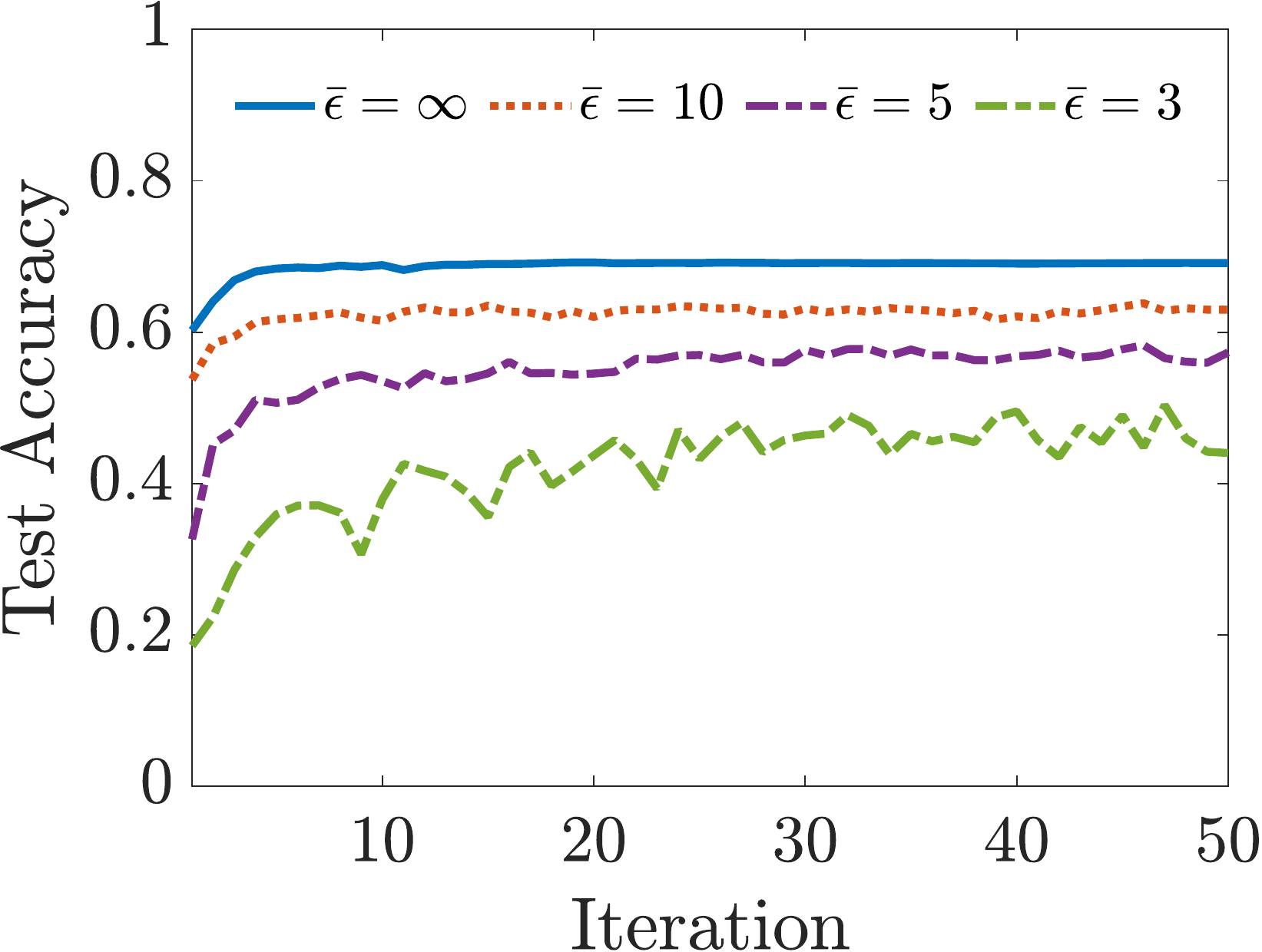}
        \includegraphics[width=\textwidth]{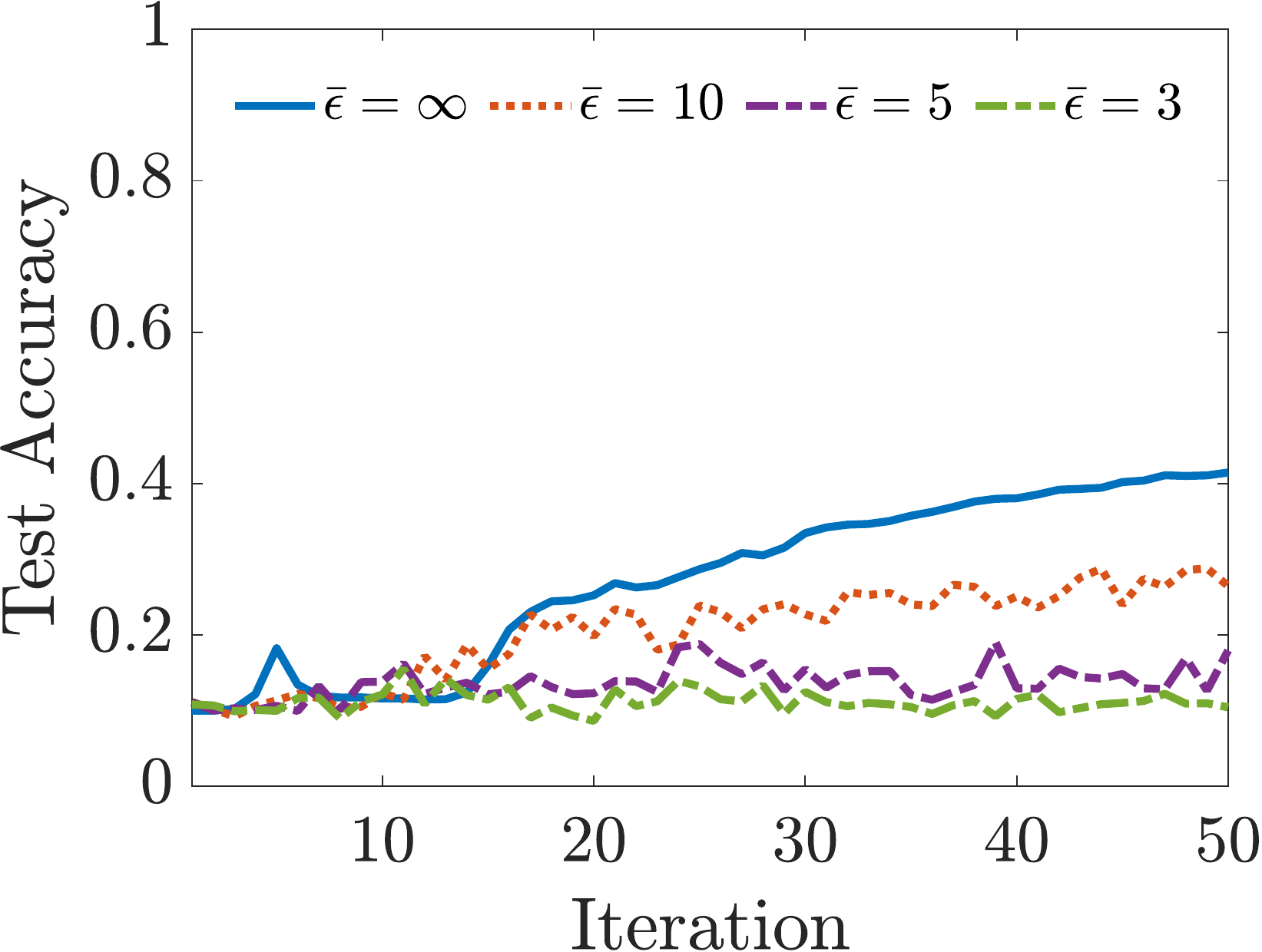}
        \includegraphics[width=\textwidth]{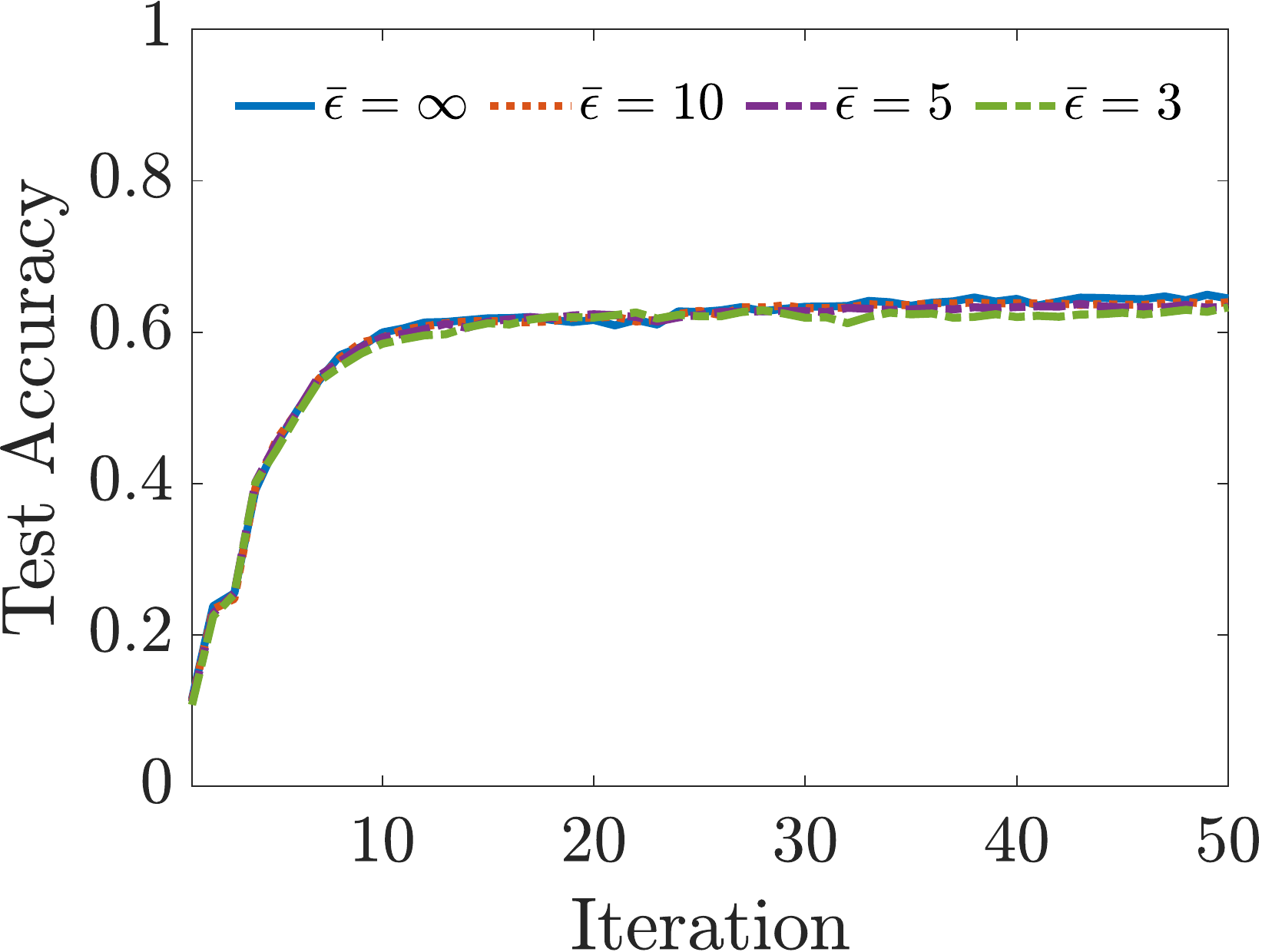}
        \caption{CIFAR10}
    \end{subfigure}
    \begin{subfigure}[b]{0.24\textwidth}
        \centering      
        \includegraphics[width=\textwidth]{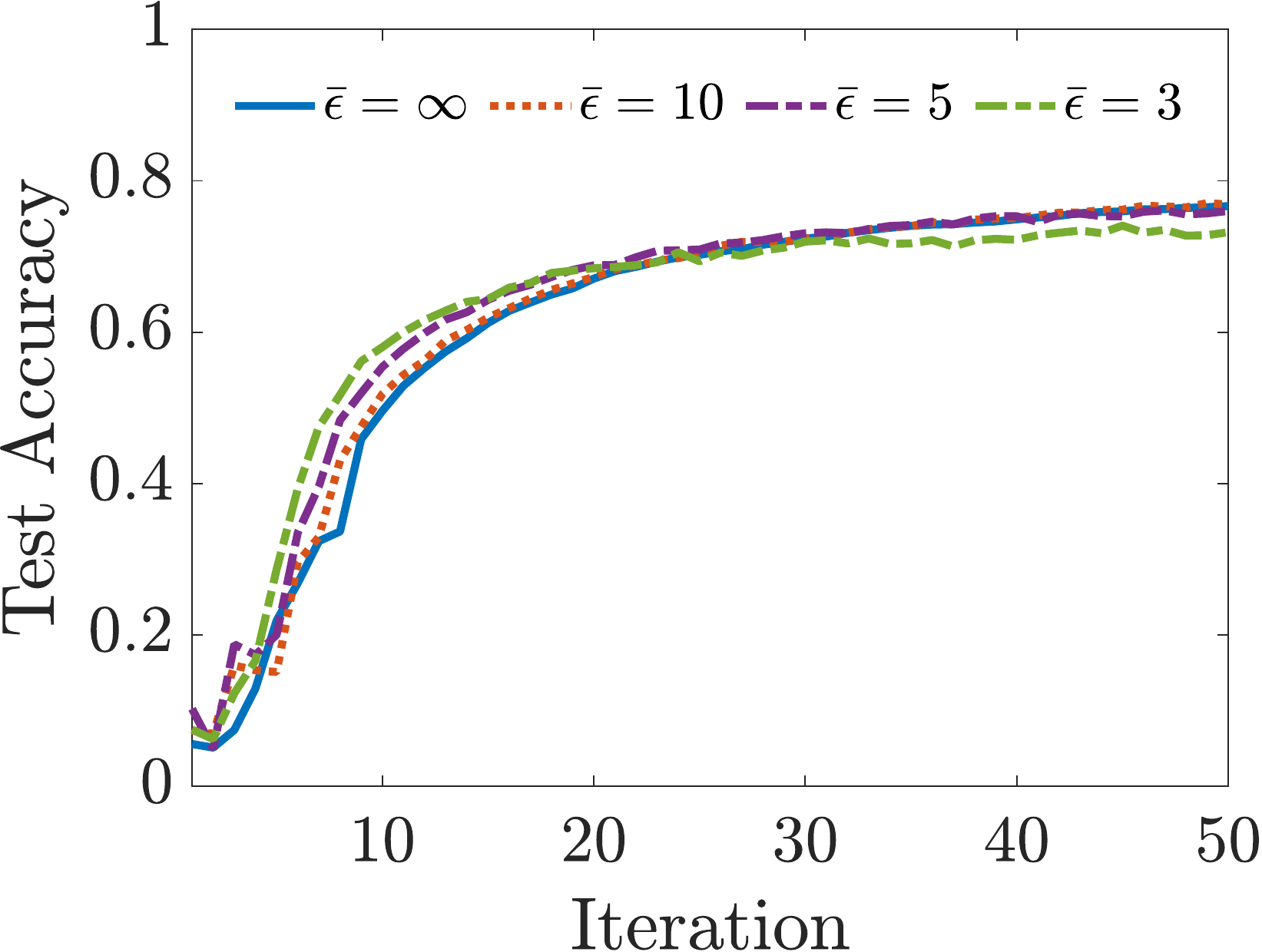}
        \includegraphics[width=\textwidth]{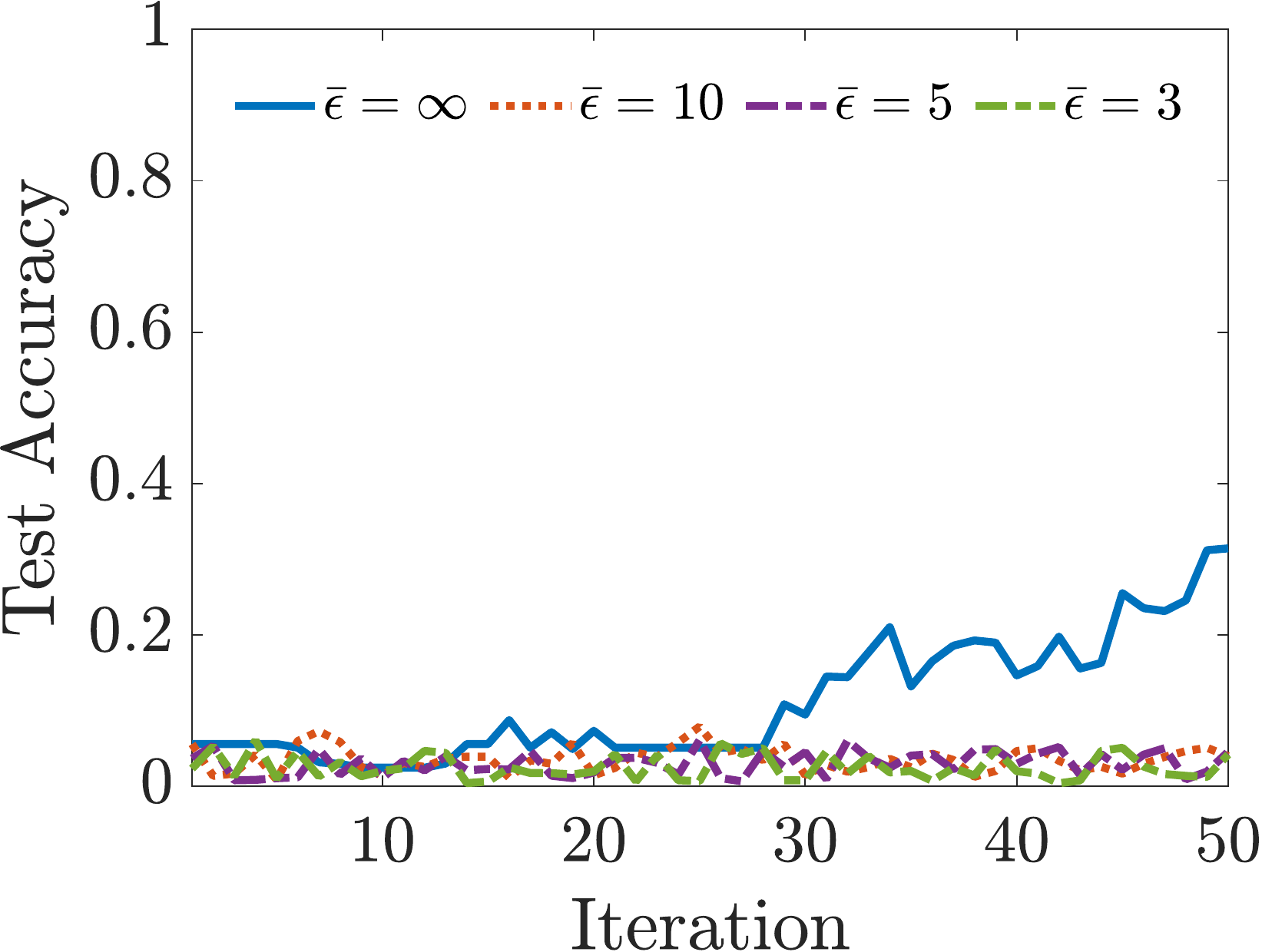}
        \includegraphics[width=\textwidth]{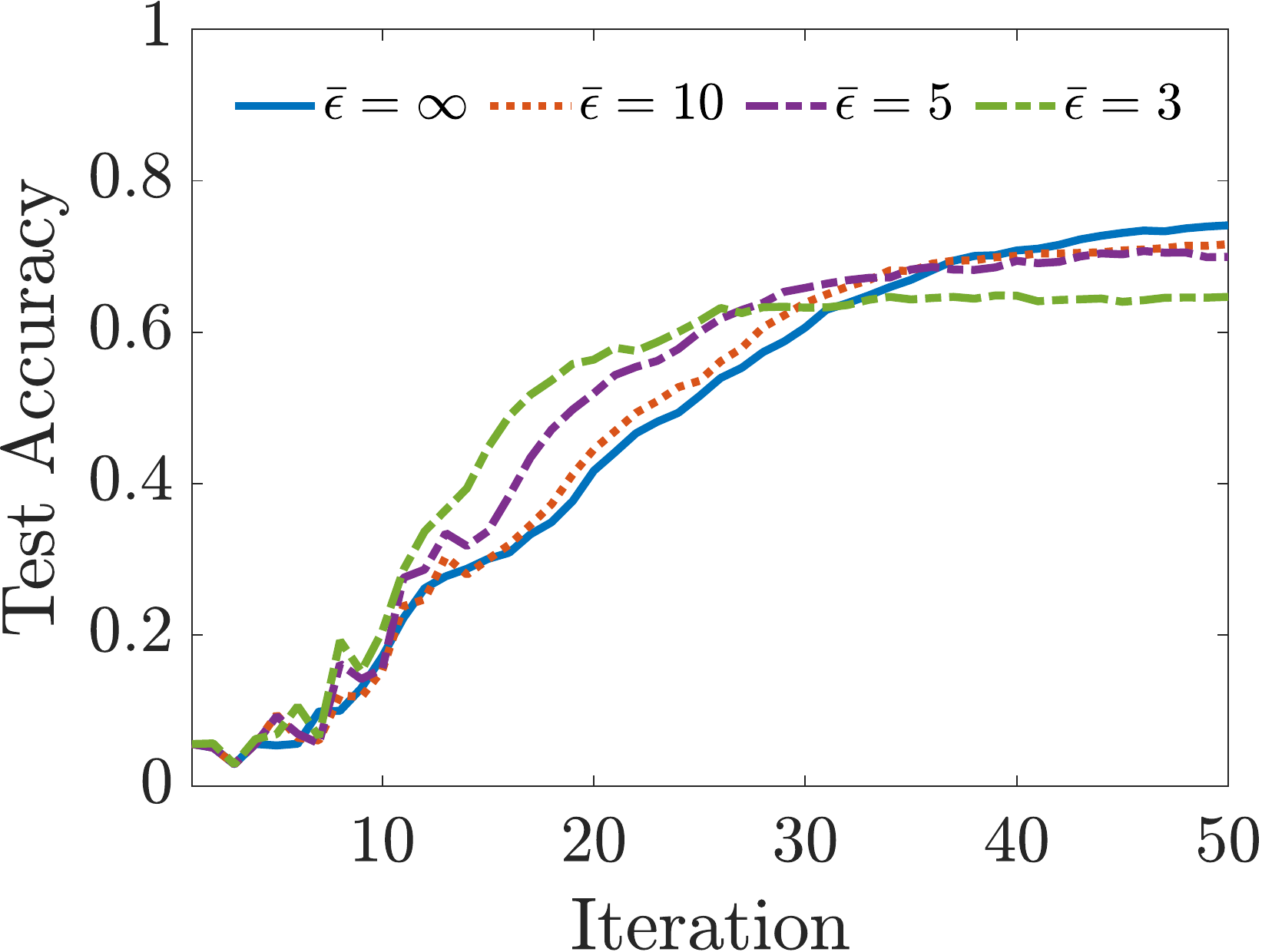}
        \caption{FEMNIST}
    \end{subfigure}  
    \begin{subfigure}[b]{0.24\textwidth}
      \centering      
      \includegraphics[width=\textwidth]{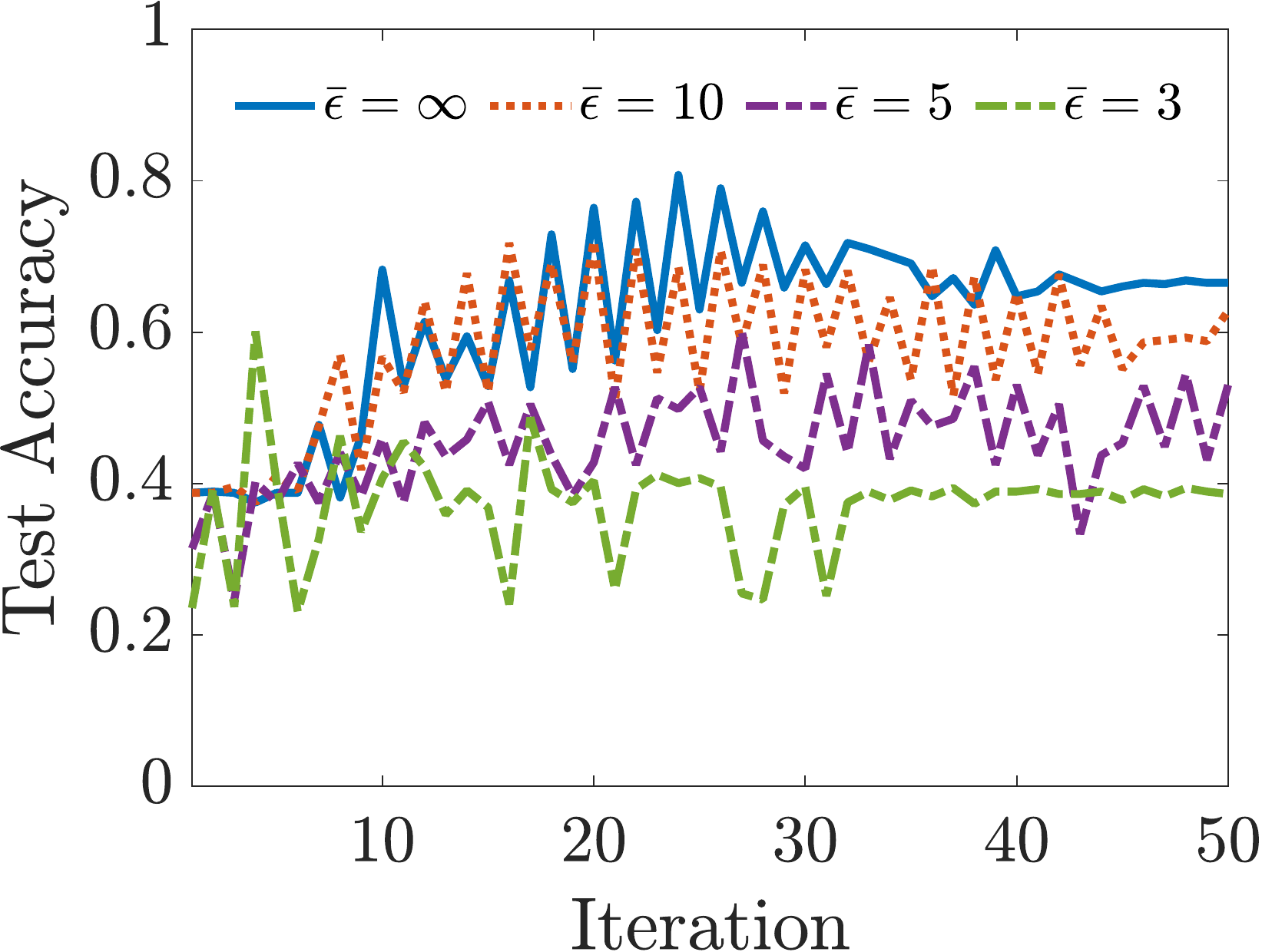}
        \includegraphics[width=\textwidth]{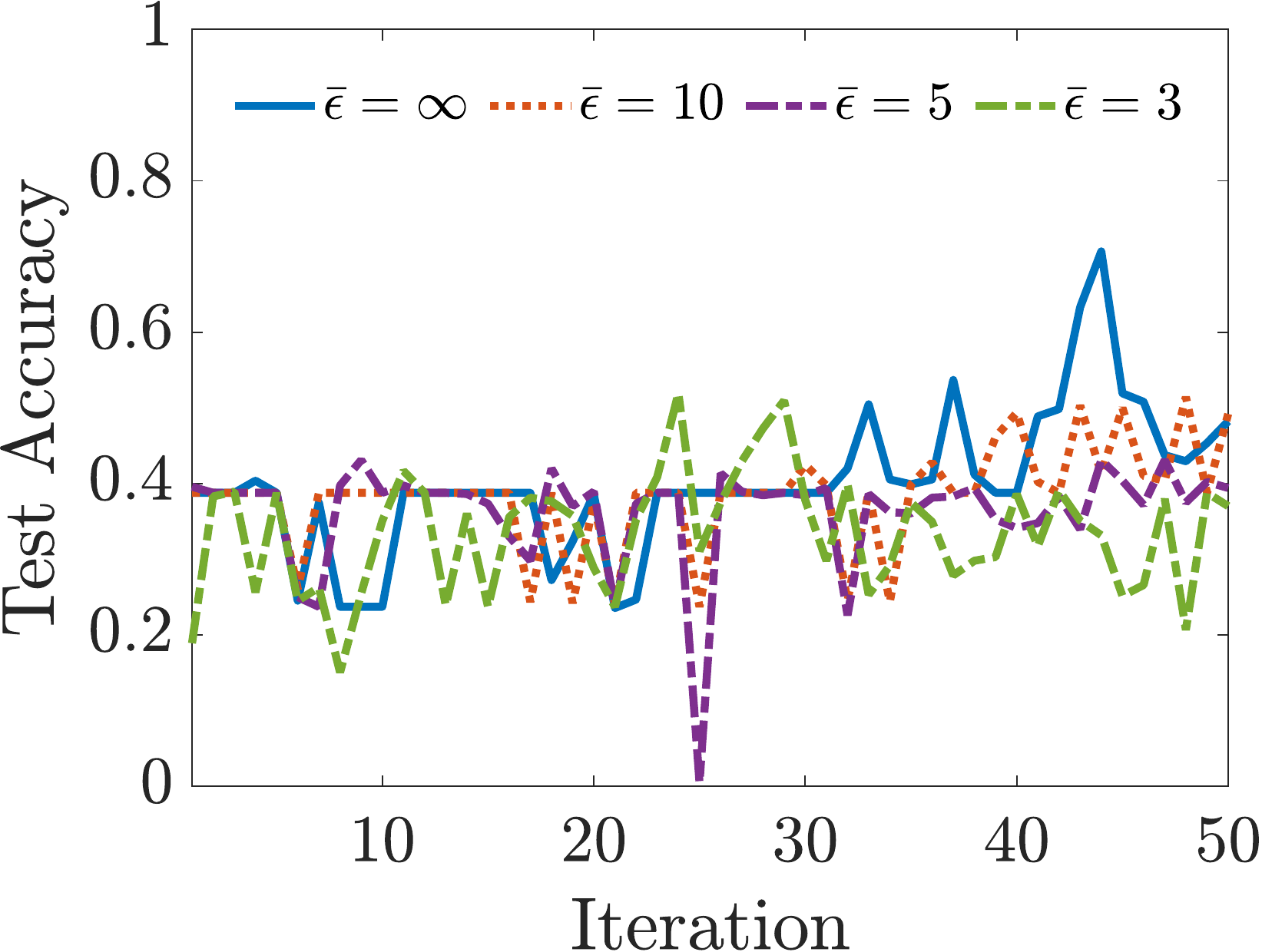}
        \includegraphics[width=\textwidth]{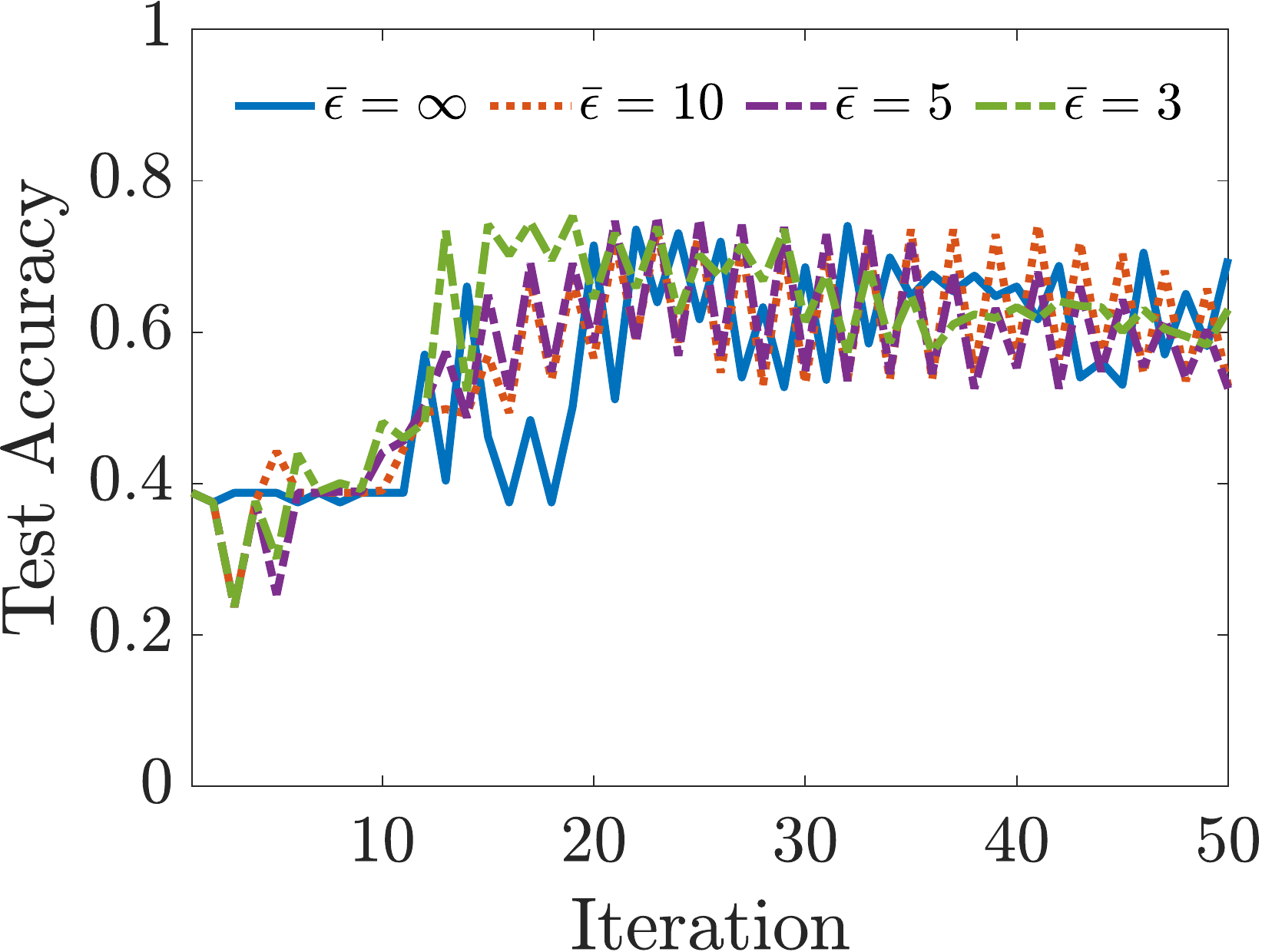}
      \caption{CoronaHack}
  \end{subfigure}  
    \caption{Test accuracy under various $\bar{\epsilon} \in \{3, 5, 10, \infty \}$ provided by FedAvg (1st row), ICEADMM (2nd row), and IIADMM (3rd row) for various datasets} 
    \label{fig:Test_accuracy}
\end{figure*}

\subsection{Experimental settings}

Our demonstration uses four datasets: MNIST, CIFAR10, FEMNIST, and CoronaHack.
The MNIST and CIFAR10 data are available from Torchvision 0.11.
The FEMNIST and CoronaHack datasets are available from the LEAF~\cite{caldas2018leaf} and kaggle~\footnote{The CoronaHack dataset has been obtained from kaggle, available at \url{https://www.kaggle.com/praveengovi/coronahack-chest-xraydataset/version/3}.} projects, respectively.
For MNIST, CIFAR10, and CoronaHack, we split the entire training datasets into four, each of which represents a client's dataset. 
The FEMNIST datasets are preprocessed to sample 5\% of the entire 805,263 data points in a non-i.i.d. (independent and identically distributed) manner, resulting in 36,699 training and 4,176 testing samples distributed over 203 clients.

We use the convolutional neural network model, consisting of two 2D convolution layers, a 2D max pooling layer, the elementwise rectified linear unit function, and two layers of linear transformation.
We note that the goal of this demonstration is not to find the best neural network architecture for each test instance or to achieve a good learning performance.

To demonstrate the differential privacy, we add random noise extracted from a Laplace distribution with zero mean and a scale parameter $b = \bar{\Delta} / \bar{\epsilon}$ to local model parameters before sending them to a central server.
Note that $\bar{\Delta}$ is a sensitivity of the local model parameters computed automatically based on the dataset and algorithm chosen in \texttt{APPFL}; therefore $\bar{\epsilon}$, from the definition of $\bar{\epsilon}$-DP in \eqref{DP_def}, controls the privacy level of the FL algorithms in \texttt{APPFL}.


\subsection{Demonstration of different PPFL algorithms}

In this subsection we demonstrate and discuss the numerical behavior of our new algorithm IIADMM (Section~\ref{sec:algorithms}), as compared with the behavior of two existing algorithms: FedAvg and ICEADMM.
We set $L=10$ local updates and $T=50$ iterations (i.e., communication rounds) for all FL algorithms.
Also, we set each batch of the local training data to have at most $64$ data points for local updates in FedAvg and IIADMM. 
Note that
(i) the SGD with momentum \cite{qian1999momentum} is utilized for FedAvg and 
(ii) all data points are used for calculating a gradient in ICEADMM as implemented in \cite{zhou2021communication}.

Experiments were run on Swing, a 6-node GPU computing cluster at Argonne National Laboratory. Each node has 8 NVIDIA A100 40GB GPUs and 128 CPU cores.
For all the algorithms, we use 1 GPU for a central server computation (i.e., global update) and 4 GPUs for clients' computation (i.e., local update).

Figure \ref{fig:Test_accuracy} displays testing accuracy resulting from the FL algorithms under the changes of privacy parameter $\bar{\epsilon} \in \{3,5,10,\infty\}$, where decreasing $\bar{\epsilon}$ ensures stronger data privacy and $\bar{\epsilon}=\infty$ represents a non-private setting.
For all FL algorithms, the results show that the test accuracy decreases as $\bar{\epsilon}$ decreases, which is a well-known trade-off between learning performance and data privacy.
\re{As compared with ICEADMM, our algorithm IIADMM provides better test accuracy in all datasets considered. This result implies the ineffectiveness of multiple local dual updates in ICEADMM as IIADMM conducts multiple local primal updates only.
As compared with FedAvg, IIADMM provides better test accuracy for most datasets (e.g., MNIST, CIFAR10, and CoronaHack) when $\bar{\epsilon}$ is smaller. This result partly demonstrates the effectiveness of the proximal term in \eqref{IADMM-2} that mitigates the negative impact of random noises generated for data privacy on the learning performance.}

We note that the magnitude of noise generated for ensuring $\bar{\epsilon}$-DP may vary over FL algorithms because it also depends on the sensitivity $\bar{\Delta}$, which varies over FL algorithms.
For example, the sensitivity in FedAvg depends on the learning rate (or step size), whereas the sensitivity in IIADMM depends on the penalty and proximity parameters, namely, $\rho^t$ and $\zeta^t$ in \eqref{IADMM-2}.
Since these parameters affect not only the data privacy but also the learning performance, they should be fine-tuned.
As part of future work, we plan to utilize both primal and dual information to further improve the performance of IIADMM.

\subsection{Scaling results of PPFL simulations on Summit}
\label{sec:scaling-mpi}

\begin{figure}[!htp]
    \centering
    \begin{subfigure}[b]{0.45\textwidth}
        \centering
        \begin{tikzpicture}
        \begin{loglogaxis}[
            width=\textwidth,
            xlabel={\# MPI processes for clients},
            ylabel={Speedup},
            xtick={5,11,24,50,101,203},
            xticklabels={5,11,24,50,101,203},
            xmajorgrids=true,
            xminorgrids=true,
            ymajorgrids=true,
            yminorgrids=true,
            ylabel near ticks,
            xlabel near ticks,
            log base y={2},
            grid style={line width=.1pt, draw=gray!30},
            legend style={legend pos=north west, cells={anchor=west}},
        ]
        \addplot[
            color=black,
            very thick,
        ] coordinates {
            (5,1)(11,11/5)(24,24/5)(50,50/5)(101,101/5)(203,203/5)
        };
        \addlegendentry{Ideal}
        
        \addplot[
            dashed,
            color=black,
            mark=square,
            mark options={scale=2,solid},
        ] coordinates {
            (5,1)(11,6.84/3.84)(24,6.84/2.02)(50,6.84/1.11)(101,6.84/0.72)(203,6.84/0.50)
        };
        \addlegendentry{APPFL}
        \end{loglogaxis}
        \end{tikzpicture}
        \caption{Strong scaling of local updates}
        \label{fig:exp-strong-scaling}
    \end{subfigure}\\\vspace{2mm}
    \begin{subfigure}{0.45\textwidth}
        \centering
        \begin{tikzpicture}
        \begin{axis}[
            width={\textwidth},
            ybar stacked,
            xlabel={\# MPI processes for clients},
            ylabel={Percentage of gather (\%)},
            xmajorgrids=true,
            xminorgrids=true,
            ymajorgrids=true,
            yminorgrids=true,
            ylabel near ticks,
            xlabel near ticks,
            xtick={1,2,3,4,5,6},
            xticklabels={5,11,24,50,101,203},
            grid style={line width=.1pt, draw=gray!30},
        ]
        \addplot coordinates {
            (1,3.60)(2,4.01)(3,7.26)(4,10.90)(5,16.55)(6,28.01)
        };
        \end{axis}
        \end{tikzpicture}
        \caption{Percentage of {\tt MPI.gather()} in local update time}
        \label{fig:exp-mpi}
    \end{subfigure}
    \caption{Scaling results of \APPFL{} on FEMNIST dataset}
    \label{fig:exp-scaling}
\end{figure}
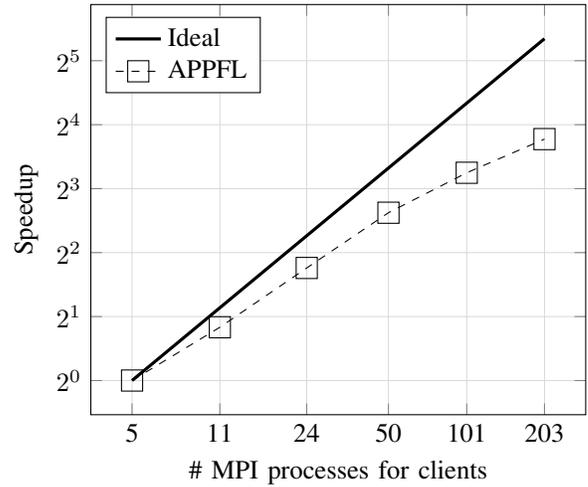
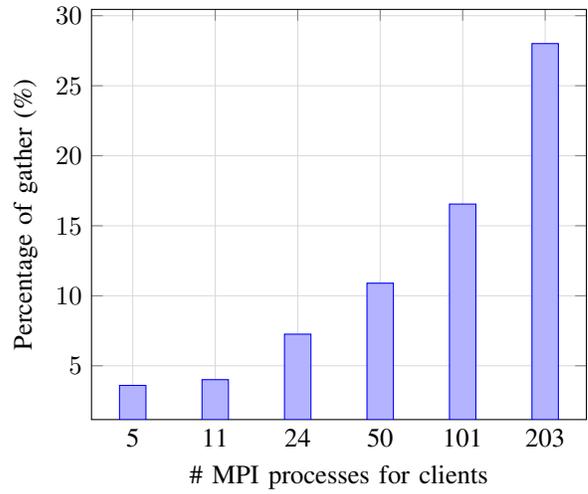

To demonstrate the scalability of PPFL over a large number of clients, we measure the average time (computation + communication) for clients' local updates over various numbers of MPI processes in \APPFL{} in a cluster.
Our experiment setup may be considered as an ideal situation where the communication efficiency of an FL is maximized via the use of InfiniBand and RDMA (remote direct memory access), which provides low latency and no extra copies of data.
As we will show later in this section, even under such an ideal situation, we observe the significance of communication efficiency in overall computation time of an FL as we increase the number of clients.

Our experiments werr performed on the FEMNIST dataset on the Summit supercomputer at Oak Ridge National Laboratory. Each compute node has 6 Nvidia V100 GPUs.
A total of $203$ clients are equally divided into a number of MPI processes, and each MPI process is assigned to a dedicated GPU.
One MPI process is reserved for a server to perform its global update.
For communications between a server and clients for local updates, we use a collective communication protocol, \texttt{MPI.gather()}, which is configured to use an RDMA technology for a direct data transfer between GPUs.
Of the 50 communication rounds, we take the average of the time over the last 49 rounds by excluding the time for the first round, which includes the compile time of the Python code.

Figure~\ref{fig:exp-strong-scaling} shows the strong-scaling results of \APPFL{} with an increasing number of MPI processes.
In the figure, the ideal plot is a reference having a perfect scaling.
\APPFL{} shows almost perfect scaling with a smaller number of MPI processes; however, its speedup decreases as we increase the number of MPI processes.
This deterioration is mainly because of the relative increase in its communication time via MPI.

More specifically, Figure~\ref{fig:exp-mpi} presents the percentage\footnote{The percentage of the communication time of each MPI process is computed by 100 $\times$ (time for {\tt MPI.gather()} / (time for {\tt MPI.gather()} $+$ compute time for local model updates)).} of MPI communication time by {\tt MPI.gather()} in the total elapsed time for local model updates for each MPI setting, which has been computed by averaging the percentages of all MPI processes.
From the results we observe that the MPI communication time does not scale as well as the pure compute time for local model updates.
While the size of data to send has decreased by more than a factor of 40 (5 vs 203 MPI processes), its communication time has decreased only by a factor of 8.
In contrast, the compute time shows perfect scaling.

Our experimental results indicate that communication efficiency may significantly affect the overall computational performance of FL as we increase the number of clients.
We plan to investigate ways to mitigate these adverse effects of communication time on training efficiency when we employ a large number of clients, such as an asynchronous update scheme and a different training scheme (e.g., dynamically controlling  the number of local updates).

\subsection{Simulations of PPFL with gRPC}\label{sec:gRPC}

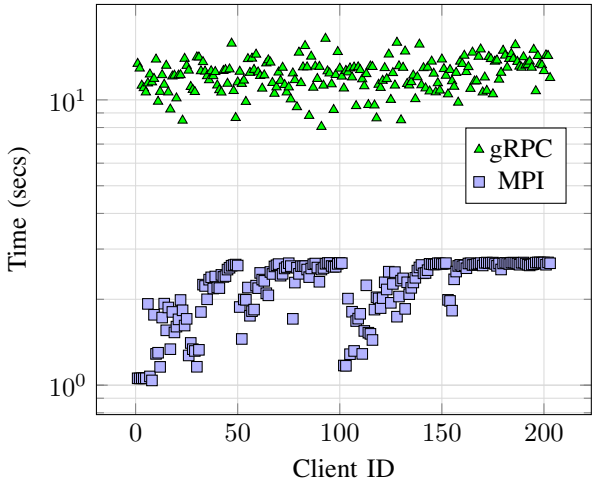
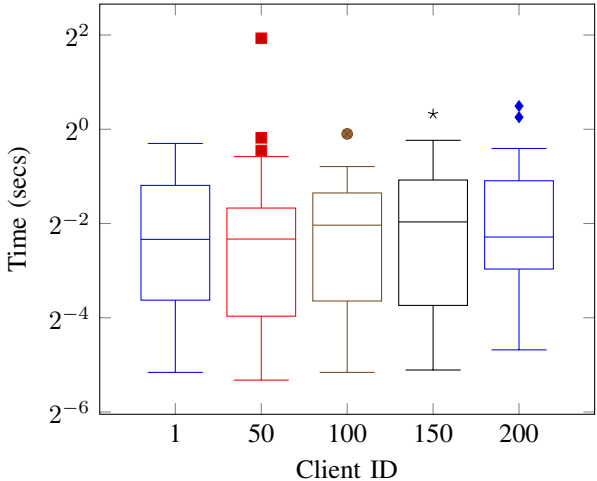
\begin{figure}[!htp]
    \centering
    \begin{subfigure}[b]{0.45\textwidth}
        \begin{tikzpicture}
            \begin{axis}[
                width={\textwidth},
                legend style={at={(0.95,0.7)}},
                xlabel={Client ID},
                ylabel={Time (secs)},
                grid style={line width=.1pt, draw=gray!30},
                ymode=log,
                xmajorgrids=true,
                xminorgrids=true,
                ymajorgrids=true,
                yminorgrids=true,
                ylabel near ticks,
                xlabel near ticks,
            ]
            
            \addplot[only marks,mark=triangle*,mark options={fill=green}] table[x=ClientID,y=Send]{./Figures/grpc.dat};
            \addlegendentry{gRPC}
            
            \addplot[only marks,mark=square*,mark options={fill=blue!30}] table[x=ClientID,y=Gather]{./Figures/mpi.dat};
            \addlegendentry{MPI}
            \end{axis}
        \end{tikzpicture}
        \caption{Cumulative communication time over 49 rounds}
        \label{fig:grpc-cumul}
    \end{subfigure}\\\vspace{2mm}
    \begin{subfigure}[b]{0.45\textwidth}
        \begin{tikzpicture}
            \begin{axis}[
                width={\textwidth},
                boxplot/draw direction=y,
                xtick={1,2,3,4,5},
                xticklabels={1,50,100,150,200},
                ymode=log,
                ylabel={Time (secs)},
                xlabel={Client ID},
                log base y={2},
                ylabel near ticks,
                xlabel near ticks,                
            ]
            
            \addplot+[boxplot] table[y index=2]{./Figures/grpc_1.dat};
            \addplot+[boxplot] table[y index=2]{./Figures/grpc_50.dat};
            \addplot+[boxplot] table[y index=2]{./Figures/grpc_100.dat};
            \addplot+[boxplot] table[y index=2]{./Figures/grpc_150.dat};
            \addplot+[boxplot] table[y index=2]{./Figures/grpc_200.dat};        
            \end{axis}
        \end{tikzpicture}
        \caption{Box plot of communication time of gRPC over 5 clients}
        \label{fig:grpc-stat}
    \end{subfigure}
    \caption{Communication times of gRPC and MPI on FEMNIST dataset}
    \label{fig:grpc}
\end{figure}

We present and discuss the impact of using gRPC on \APPFL{}, as compared with MPI, over the FEMNIST dataset.
This comparison is important in order to understand the communication efficiency of real-world FL settings because they typically involve clients remotely apart from each other with heterogeneous architectures, whereas MPI is available only on a cluster environment.
In such settings, we may not be able to exploit fast network devices and protocols such as InfiniBand and RDMA that we employed in Section~\ref{sec:scaling-mpi}.
These limitstions may result in even worse efficiency of the network communications in the total computation time than the case for MPI, as we have seen in Figure~\ref{fig:exp-mpi}.

To perform the experiments, we have a setup similar to that in Section~\ref{sec:scaling-mpi}, except that we now use gRPC for communication.
A total of 203 clients have been launched on 34 compute nodes (physically apart from one another) on the Summit cluster where each node (except for the last node) runs 6 clients with a dedicated GPU being assigned to each of them.
Although these nodes are connected via InfiniBand, our gRPC is not configured to use RDMA, so communication via gRPC may not be as efficient as with MPI, which is configured to directly transfer data between GPUs via RDMA.
Our configuration provides a more realistic network environment.

Figure~\ref{fig:grpc-cumul} presents a comparison of cumulative communication times between MPI and gRPC over 49 rounds excluding the first round since it typically involves compile time.
As we see in the figure, MPI shows up to 10 times faster communication time than does gRPC.
We think that the main reason for this performance degradation of gRPC is that (i) it performs serialization and deserialization of user-given data via protocol buffers and (ii) it involves copying data from GPUs to CPUs, in contrast to RDMA-enabled MPI where we directly transfer data between GPUs.

Another reason for such a degraded performance is that gRPC tends to show inconsistent communication time between rounds, as illustrated in Figure~\ref{fig:grpc-stat}.
We sample 5 clients with IDs  1, 5, 100, 150, and 200 and present a box plot showing quantile information of communication times over 49 rounds.
From the figure, we observe a significant difference in communication time by a factor of 30 between rounds.
This could be viewed as different communication times depending on network traffic.

Similar to the observations described in Section~\ref{sec:scaling-mpi}, we believe an asynchronous update scheme of an FL will allow us to more efficiently perform FL in the presence of this communication inefficiency.

\subsection{Impact of heterogeneous architectures}

We next discuss the impact of heterogeneous architectures with some quantitative evidence. 
While most FL studies are simulated on homogeneous computing architectures, a practical FL setting may be composed of many heterogeneous computing machines. 
Consider a cross-silo setting where one institution updates the local model on a machine with NVIDIA A100 GPUs (e.g., on Argonne's Swing) and the other institution updates the local model on a machine with NVIDIA V100 GPUs (e.g., on Oak Ridge's Summit).
This can cause a significant load imbalance between the two local updates. 
For example, the local update on one A100 GPU is faster than that on one V100 GPU by a factor of 1.64 (6.96 seconds vs. 4.24 seconds).
This implies that the heterogeneous architectures in FL will be an important factor for the design of efficient FL algorithms.

\section{Concluding Remarks and Future Work} \label{sec:conclusion}

In this paper we introduced \APPFL\, our open-source PPFL framework  that allows research communities to develop, test, and benchmark FL algorithms, data privacy techniques, and neural network architectures for decentralized data.
In addition to the implementation of existing FL algorithms, we have developed and implemented a new communication-efficient FL algorithm that significantly reduces the communication data every iteration. 
Two communication protocols, gRPC and MPI, have been implemented and numerically demonstrated with \APPFL. 
In particular, we demonstrated the communication performance of using gRPC and the scalability of distributed training with MPI on the Summit supercomputer.

Many interesting and challenging questions are being actively investigated by the FL research communities.
We conclude this paper by discussing our future technical work for \texttt{APPFL}.
\begin{enumerate}
    \item The current communication topology used in our framework is based on a client-server architecture, which may suffer from load imbalance in local computations. We plan to implement the asynchronous updates of an FL model in our framework. We will also develop decentralized privacy-preserving algorithms that allow the neighboring communication without the central server for learning.
    \item We will enhance the learning performance of IIADMM by adaptively updating algorithm parameters such as penalty $\rho^t$ and proximity $\zeta^t$. In addition to existing techniques, ML approaches (e.g., reinforcement learning~\cite{zeng2021reinforcement}) can be used for updating such parameters.
    \item Computation of the sensitivity parameter $\bar{\Delta}$ used in Section~\ref{sec:demo} is key to achieving greater learning performance while preserving data privacy. We will develop efficient algorithms to compute the scale parameter for differential privacy. 
    \item To better understand the communication bottleneck among devices (vs. nodes on a cluster), we will test our framework with large-scale deep neural network models that require a large amount of data transfer between a server and clients.
\end{enumerate}

\section*{Acknowledgment}
We gratefully acknowledge the computing resources provided on Swing, a high-performance computing cluster operated by the Laboratory Computing Resource Center at Argonne National Laboratory.
This research also used resources of the Oak Ridge Leadership Computing Facility at the Oak Ridge National Laboratory, which is supported by the Office of Science of the U.S. Department of Energy under Contract No. DE-AC05-00OR22725.

\bibliographystyle{IEEEtran}
\bibliography{References.bib}

\end{document}